\def\year{2021}\relax
\documentclass[letterpaper]{article} 
\usepackage{aaai21}  
\usepackage{times}  
\usepackage{helvet} 
\usepackage{courier}  
\usepackage[hyphens]{url}  
\usepackage{graphicx} 
\usepackage{graphicx}  
\usepackage{natbib}  
\usepackage{caption} 
\usepackage{amsmath}
\usepackage{amsthm}
\usepackage{amsfonts}
\usepackage{amssymb}
\usepackage{booktabs}

\usepackage{multirow}
\usepackage{microtype}
\usepackage{nicefrac}
\usepackage{empheq}
\usepackage{subfig}
\usepackage[switch]{lineno}

\newcommand{\eat}[1]{}
\newcommand{\LeafInfluence}{\emph{LeafInfluence}}

\title{TREX: Tree-Ensemble Representer-Point Explanations}

\author {
  Jonathan Brophy, \textsuperscript{\rm 1}
  Daniel Lowd \textsuperscript{\rm 1} \\
}
\affiliations {
    \textsuperscript{\rm 1} University of Oregon \\
    \{jbrophy, lowd\}@cs.uoregon.edu
}

\begin{document}

\maketitle

\begin{abstract}
How can we identify the training examples that contribute most to the prediction of a tree ensemble? In this paper, we introduce TREX, an explanation system that provides instance-attribution explanations for tree ensembles, such as random forests and gradient boosted trees. TREX builds on the representer point framework previously developed for explaining deep neural networks. Since tree ensembles are non-differentiable, we define a kernel that captures the structure of the specific tree ensemble. By using this kernel in kernel logistic regression or a support vector machine, TREX builds a surrogate model that approximates the original tree ensemble. The weights in the kernel expansion of the surrogate model are used to define the global or local importance of each training example.

Our experiments show that TREX's surrogate model accurately approximates the tree ensemble; its global importance weights are more effective in dataset debugging than the previous state-of-the-art; its explanations identify the most influential samples better than alternative methods under the remove and retrain evaluation framework; it runs orders of magnitude faster than alternative methods; and its local explanations can identify and explain errors due to domain mismatch.
\end{abstract}

\section{Introduction}

Tree ensembles, including random forests~\cite{breiman2001random} and gradient boosted trees~\cite{friedman2001greedy}, remain one of the most effective \eat{machine learning} approaches to classification. In recent years, their popularity has only grown, as shown by the increasing number of gradient boosting frameworks, including XGBoost~\cite{chen2016xgboost}, LightGBM~\cite{ke2017lightgbm}, and CatBoost~\cite{prokhorenkova2018catboost}.

As performant as tree ensembles are, their complexity and scale can make them 
difficult to understand, 
leading to lower trust and decreased use, 
especially in problem domains where decisions can have major impacts~(e.g. health care, autonomous vehicles, etc.). By understanding how these models make predictions at a deeper level, we can expose deficiencies in the model or the data they are trained on,
leading to more accurate and trustworthy models.

In this paper, we focus on understanding models through
{\em instance attribution} explanations, which identify examples in the training data that have the greatest impact on a model's prediction for a given query instance. 
These explanations provide insight into both the models and the data used to train them.
Instance attribution methods can also be used in conjunction with other explanation approaches, such as feature-attribution methods~\cite{lundberg2017unified,ribeiro2016should}.

The two main approaches to instance attribution are influence functions~\cite{belsley2005regression}, which approximate the marginal effect of increasing an example's weight on the final model, and representer points~\cite{yeh2018representer}, which build on representer theorems~\cite{scholkopf2001generalized} to approximate the target model with a weighted sum of the training points. The main limitation of influence functions is efficiency: even with heuristics and approximations, computing the influence of each training point is extremely expensive, as we show in our experiments. Previous work on representer point methods is 
specific to deep learning models and requires a suitable differentiable loss function to compute the ``representer values'' (weights) of the training-instance activations. Prototype selection methods also summarize a dataset or model with a set of training examples, but these methods do not provide local explanations for individual predictions~\cite{bien2011prototype,kim2016examples,gurumoorthy2019efficient}.


Tree ensembles are non-differentiable due to the step functions created by feature splits in each tree. To adapt the representer point framework to tree ensembles, we introduce TREX~(\textbf{T}ree-ensemble \textbf{R}epresenter-point \textbf{EX}planations). TREX leverages the structure of the trees to build a tree ensemble kernel, which acts as a similarity measure between data points. Then, it trains a kernelized model on this new kernel representation by solving the dual problem to obtain weights for the training instances. We are then able to decompose any prediction as a linear combination of these training instances, resulting in an instance-attribution explanation of positive and negative training points. We show that TREX is not only able to aid in dataset debugging and model understanding, but is also more scalable than previous methods.

Our contributions are as follows:
\begin{enumerate}
\item We define a new kernel for tree ensembles, LeafOutput, based on the path through and the numerical output of each tree in the ensemble.
\item We introduce TREX, a method for generating global and local explanations for tree ensembles by using a kernelized surrogate model within the representer point framework.
\item We evaluate TREX on four benchmark datasets, demonstrating that: (1) the surrogate models accurately approximate the original tree ensembles; (2) in a data cleaning setting, TREX identifies noisy instances better than the previous state-of-the-art; (3) in a remove-and-retrain setting, TREX identifies the training examples most important for good performance; (4) TREX is orders of magnitude faster than other methods; (5) TREX's local explanations can identify and explain errors due to domain mismatch.
\end{enumerate}

\section{Background}

There are a number works in the interpretable machine learning literature~\cite{adadi2018peeking,miller2019explanation} that analyze the importance of different features from a global model perspective~\cite{kazemitabar2017variable} as well as a local perspective on specific model predictions; these typically pertain to feature-attribution methods~\cite{treeinterpreter,ribeiro2016should,lundberg2017unified,lundberg2018consistent} or counterfactuals~\cite{guidotti2019factual}. 

\subsubsection{Sample-Based Explanations} Although these works give insight into which features are important for a given model /  prediction, they do not provide information about specific important or influential training samples. Thus, a growing body of work focuses on identifying the potentially influential training samples that can help automate the process of debugging noisy datasets and better understand model behavior, among others applications~\cite{koh2017understanding}. For example, prototype selection methods~\cite{bien2011prototype,kim2016examples,gurumoorthy2019efficient} provide a global perspective of a dataset by analyzing the training examples in high~(prototypes) and low~(criticisms) density regions of the input space. However, prototype selection methods do not provide any information about \emph{which} training examples are most influential for a \emph{given} model prediction, also known as local explanations; to address this issue, we turn to instance-attribution explanations.

\subsubsection{Instance-Attribution Explanations}

A naive but intractable approach to evaluating the impact of each individual training sample on a given prediction involves leave-one-out retraining~\cite{belsley2005regression} for every training instance. To avoid this problem, \citet{koh2017understanding} derived an approximation of the influence functions framework from classical statistics~\cite{cook1980characterizations} for deep learning models; their approach allows one to compute the influence of each training instance on a given model prediction without having to retrain the model from scratch. However, their approach is specific to deep learning models and requires the computation of the Hessian vector product and first-order derivatives of the loss function. This can be quite an expensive operation, thus~\citet{yeh2018representer} introduce the representer point framework for deep neural networks, a method based on the representer theorem~\cite{scholkopf2001generalized} as a way to efficiently decompose the pre-activation predictions of a neural network into a linear combination of the training samples.

\subsubsection{Instance-Attribution Explanations for Tree Ensembles}

Since influence functions cannot be directly applied to decision trees, \citet{pmlr-v80-sharchilev18a} introduce~\LeafInfluence, an extension of influence functions to gradient boosted decision trees. Their approach considers the tree-ensemble structure to be fixed, allowing them to analyze the changes in leaf values with respect to the weights of the training samples. However, their approach is still based on the influence functions framework designed by~\citet{koh2017understanding}, and thus requires expensive first and second-order derivative computations to find the influential training samples.

\citet{plumb2018model} take a different approach with MAPLE, a model-agnostic supervised local explainer. MAPLE fits a regression forest to the outputs of a black-box model, and then uses a feature importance selector called DSTUMP~\cite{kazemitabar2017variable} to select the most important features. When an explanation is desirable, MAPLE uses SILO~\cite{bloniarz2016supervised}, a local linear modeling technique that uses random forests to identify supervised neighbors~\cite{davies2014random,he2014practical}, to generate a prediction. Specifically, given an instance to predict $x_t$, SILO generates a \emph{local training distribution} based on how often a training instance $x_i$ ends at the same terminal node as $x_t$. This distribution is then used to fit a weighted linear regression model, whose prediction approximates $E(y|x_t)$. MAPLE uses the coefficients of this linear model as a feature-attribution explanation, similar to LIME~\cite{ribeiro2016should}; MAPLE also uses the resultant local training distribution as an instance-attribution explanation, representing the most important samples to the prediction of $x_t$.

Instead of fitting a separate weighted linear model for each prediction, our approach trains a kernelized model that globally approximates the tree ensemble; we can use this model to get a global perspective of which training samples are influential to the model as well as provide local explanations as to which samples are the most influential for a given test instance or a set of test instances. We build this kernel model by using a new supervised tree-ensemble kernel based on the leaf outputs of the trees in the ensemble.

\section{TREX}

In this section, we present TREX (\textbf{T}ree-ensemble \textbf{R}epresenter-point \textbf{EX}planations), an extension of the representer point framework~\cite{yeh2018representer}, initially designed for deep learning models, to non-differentiable tree ensembles such as gradient boosted trees.

\subsection{Preliminaries}

We assume an {\em instance space} $\mathcal{X}$ defined over $d$ features, and denote an instance $i$ with feature $j$ as $x_{i, j}$.\eat{ $\{x_{i, 1}, x_{i, 2}, \ldots, x_{i, d}\}$, where $x \in \mathcal{X}$} For simplicity, we assume real-valued attributes and a binary label. In binary classification, our goal is to find a function $f: \mathcal{X} \rightarrow \{-1, +1\}$ that maps each instance to either the positive ($+1$) or negative ($-1$) class.

A {\em decision tree} is a tree-structured model where each leaf is associated with a categorical or real-valued prediction, each internal node is associated with an attribute $x_{i, j}$, and its outgoing branches define a partition over the attribute's values. Given an instance $x_t \in \mathcal{X}$, the prediction of a decision tree can be found by traversing the tree, starting at the root and following the branches consistent with the attribute values in $x_t$. The traversal ends in a single leaf node, and the prediction of the decision tree is equal to the value of the leaf node. For binary classification, positive leaf values denote the positive class and non-positive values denote the negative class.

A {\em tree ensemble} is a set of decision trees, each defined over the instance space $\mathcal{X}$. There are many different methods for learning tree ensembles from data, most using some form of bagging or boosting in order to induce a collection of diverse trees that work well together. Given an instance $x_t \in \mathcal{X}$, the prediction of the tree ensemble is an aggregation of the predictions of all trees in the set; typically, random forests take the average among the predictions of the trees in the forest, while gradient boosted decision trees (GBDTs) sum up the predictions from each tree.

See Figure~\ref{fig:tree_kernel} for an example of a GBDT tree ensemble containing two trees, each defined over three attributes. As depicted, the left branch of each split is associated with the value 0 and the right branch is associated with 1. For the instance $x = (1, 0, 0)$, the first tree evaluates to $5$ and the second tree evaluates to $3.8$, for a total prediction of $8.8$. Since $8.8$ is positive, the predicted label for $x$ is positive.

\subsection{Representer Point Decomposition}\label{sec:representer_point_decomposition}

Representer theorems~\cite{scholkopf2001generalized} state that the optimal solutions of many learning problems can be represented in terms of the training examples. In particular, the nonparametric representer theorem (Theorem~4 from \citet{scholkopf2001generalized}) applies to empirical risk minimization within a reproducing kernel Hilbert space (RKHS). This covers a wide range of linear and kernelized machine learning methods. 

Representer theorems provide a natural way to explain classifiers and their predictions in terms of the training instances. Given a representation in the following form,
\begin{align}\label{eq:kernel_form}
f(\cdot) &= \sum_{i=1}^m \alpha_i y_i k(\cdot, x_i),
\end{align}
the value of each weight $\alpha_i$ describes the global contribution of training instance $x_i$ to the overall classifier $f(\cdot)$. The contribution of $x_i$ to the \emph{individual} prediction of a test instance $x_t$ is simply $\alpha_i y_i k(x_t, x_i)$, the product of the training instance weight, label, and similarity to $x_t$.

However, representer theorems do not apply directly to tree ensembles  because the space of tree ensembles (with a fixed number of trees) is not a vector space.
Nonetheless, we would like to apply the ideas behind representer point methods to tree ensembles. To do so, we need a surrogate empirical risk minimization problem in an RKHS that is analogous to the tree ensemble learning task, so that explanations for the surrogate problem are useful for explaining the original tree ensemble.

Previous work on representer point methods for deep networks keeps the many non-linear layers of the network fixed, treating them as a feature map, and focuses on explaining the final, linear layer of the network~\cite{yeh2018representer}. We adopt a similar approach with tree ensembles by defining feature maps (and corresponding kernels) that capture the learned features implicit in the tree ensemble's structure and focus on explaining how these features combine to produce an overall prediction. 

We first describe the kernels we use to represent the structure of the tree ensemble (\S\ref{sec:kernels}), then the empirical risk minimization problems we use them in (\S\ref{sec:erm}), and later evaluate the utility of this approach in generating explanations in a variety of contexts (\S\ref{sec:eval}).

\subsection{Tree-Ensemble Kernels}\label{subsec:feature_transformation}
\label{sec:kernels}

The first part of our method is to 
define a kernel that computes the similarity between pairs of data points based on how they are processed by a specific tree ensemble, $T$.
Intuitively, two data points are processed identically if they are assigned to the same leaf in each tree in the ensemble. The degree of similarity between two data points can thus be defined by comparing the specific leaf or leaf value that each data point is assigned by each tree in the ensemble. 

We define our tree ensemble kernels as dot products in an alternate feature representation defined by the feature mapping $\phi$:
$k(x_i, x_j; T) = \phi(x_i; T) \cdot \phi(x_j; T)$. Note that the kernel is parametrized by $T$, since the computation necessarily depends on the structure of the tree ensemble.
Different choices of $\phi$ emphasize different aspects of the tree ensemble structure:
\begin{itemize}
    \item \emph{LeafPath}: A tree-based kernel~\cite{bloniarz2016supervised,he2014practical,plumb2018model} where the elements in this new feature vector represent the leaves of all the trees in $T$; a value of 1 means $x_i$ traversed to that leaf, otherwise the value is 0.
    
    \eat{
    \item \emph{FeaturePath}: A slightly adapted random forest kernel~\cite{davies2014random}, this method is similar to the LeafPath approach, except the vector now represents all nodes in the tree ensemble, where 1 means $x_{i, j}$ traversed through that node, and 0 otherwise. This can capture more partial similarity between data points, when some of the same branches are taken but the final leaves are different. The resulting binary vector's length is the total number of nodes in the tree ensemble.
    }

    \item \emph{TreeOutput}: A new tree-based kernel that takes the leaf value from each tree, resulting in a vector whose length is equal to the number of trees in the ensemble. This kernel supports the fact that two instances can take different paths through a tree but still contribute to the same label.

    \item \emph{LeafOutput}: A combination of the previous two kernels, in which we take the LeafPath representation and replace each value of 1 with the actual value of the corresponding leaf.
    
\end{itemize}

\begin{figure}[ht]
\centering
\includegraphics[width=0.475\textwidth]{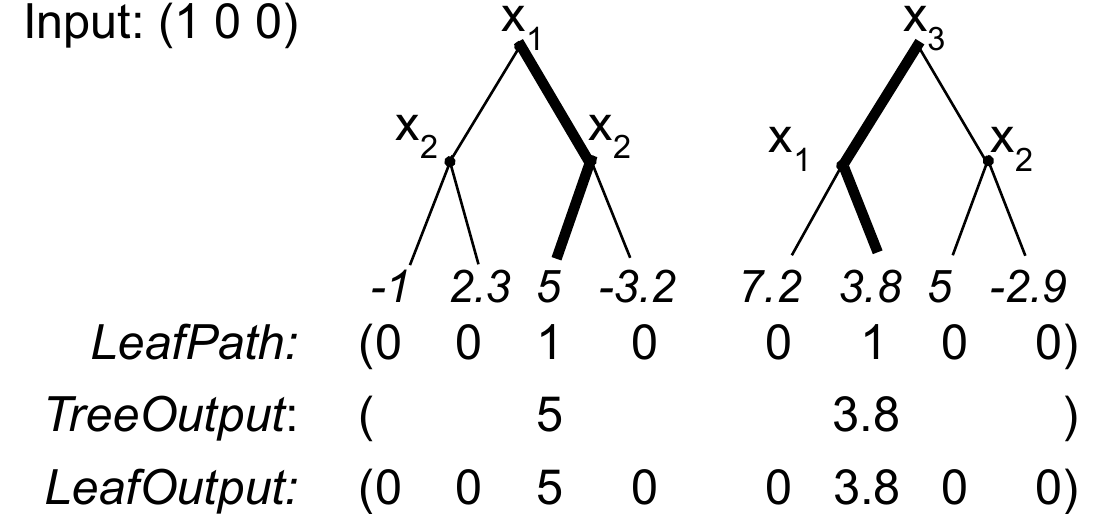}
\caption{Different transformations of a single data instance from a simple two-tree ensemble. The numbers at the leaves represent leaf values, and the lines in bold represent the paths taken through each tree in the ensemble given the input instance.}
\label{fig:tree_kernel}
\end{figure}

See Figure~\ref{fig:tree_kernel} for a simple example of each kernel's feature representation.

To be effective in generating explanations, these kernels must help approximate the tree ensemble learning problem. 
With each of the feature maps described above, the tree ensemble's prediction can be perfectly represented as a dot product with an appropriately chosen weight vector, $w^T \phi(x_i)$.
For \emph{LeafPath}, the weight vector must contain the value of each leaf.
For \emph{TreeOutput} and \emph{LeafOutput}, a weight vector of entirely 1s will suffice, summing the prediction from each tree or leaf to obtain the overall prediction from the ensemble.

\subsection{Empirical Risk Minimization}
\label{sec:erm}

The second part of our method is to combine a tree ensemble kernel with an empirical risk minimization problem. We solve these problems in the dual, obtaining a weight for each training example that denotes its relative influence in the model.

\paragraph{Kernel Logistic Regression}

Consider a tree ensemble $T$ trained on a dataset $D^*$~=~\{$x_i,y_i$\}$_1^n$ where $x_i \in \mathcal{X}$, $y_i \in \{-1, +1\}$ and $\hat{y} = \{\hat{y}_i\}_1^n$ where $\hat{y}_i \in \{-1, +1\}$ are the predicted labels from $T$ on $D^*$.
We fit a kernel logistic regression (KLR) model by optimizing the following dual objective~(\citeauthor{yu2011dual}~\citeyear{yu2011dual},~equation~(2)):
\begin{equation}\label{eq:lr_dual}
\begin{split}
    \min_{\alpha} \frac{1}{2} \alpha^T Q \alpha \sum_{i=1}^n \alpha_i + \log \alpha_i + (C - \alpha_i) \log (C - \alpha_i), \\
    \text{ s.t. } 0 \leq \alpha_i \leq C
\\
\end{split}
\end{equation}
where $C$ is a penalty parameter, $Q_{ij} = \hat{y}_i \hat{y}_j k(x_i, x_j; T)$ $\forall i, j$, and $k(\cdot, \cdot)$ is one of the kernels from the previous section. This optimization yields a value for each $\alpha_i$, representing the weight of training sample $x_i$. \citet{yeh2018representer} coin these terms ``representer values''; semantically, they represent the resistance of training instance $x_i$ to minimizing the norm of the weight matrix. 

Once we have optimized for $\alpha$, we can decompose the prediction of a new test instance $x_t$ into a weighted sum of the training samples, as per equation~(\ref{eq:kernel_form}). Thus, we can define the contribution of training instance $x_i$ to the prediction of $x_t$ as $\alpha_i \hat{y}_i k(x_i, x_t; T)$, the similarity between $x_i$ and $x_t$, weighted by $x_i$'s representer value and multiplied by the predicted label. The resulting value can be positive or negative, leading to excitatory or inhibitory examples that contribute towards or away from the predicted label of $x_t$. The result is an explanation defined in terms of all training instances since most of the values in $\alpha$ are likely to be non-zero; for a sparse-solution, we turn to support vector machines.

\paragraph{Support Vector Machine}

Following the same setup, we use an SVM~\cite{cortes1995support}
as our empirical risk minimizer. Again, we optimize the dual objective~(\citeauthor{yu2011dual}~\citeyear{yu2011dual},~equation~(4)) to obtain instance weights $\alpha$:
\begin{align}\label{eq:svm_dual}
  \min_{\alpha}\frac{1}{2}\alpha^T Q \alpha - \sum_{i=1}^n \alpha_i,
  \text{ s.t. } 0 \leq \alpha_i \leq C.
\end{align}
Since the support vectors of an SVM are the only instances with non-zero weights, the resulting explanation is sparse; this can be beneficial as the computation needed to explain a new test instance may be significantly reduced.

\section{Experiments}
\label{sec:eval}

We conduct a range of experiments to evaluate the effectiveness of TREX in a variety of contexts.

\paragraph{Datasets and Framework}

Our evaluation uses the following datasets, several of which are the same as in previous work~\cite{pmlr-v80-sharchilev18a}: \texttt{Churn}~($n=7,043$, $d=19$)~\cite{churn}, which tracks customer retention;
\texttt{Amazon}~($n=32,769$, $d=9$)~\cite{amazon}, where the task is to predict employee access for certain tasks;
\texttt{Adult}~($n=48,842$, $d=14$)~\cite{Dua:2019}, a dataset containing information about personal incomes; and
\texttt{Census}~($n=299,285$, $d=41$)~\cite{Dua:2019}, a population survey dataset collected by the U.S. Census Bureau. The Churn dataset does not have a predefined train/test split, so we randomly select 20\% to use as a test set.

Our experiments use CatBoost~\cite{prokhorenkova2018catboost}, an open source implementation of gradient boosted trees\footnote{We chose CatBoost over other implementations primarily because the LeafInfluence baseline~\cite{pmlr-v80-sharchilev18a} is implemented with CatBoost, making it easier to perform a fair comparison. We have also applied TREX to LightGBM~\cite{ke2017lightgbm}, with similar results.}. When training TREX with logistic regression, we use liblinear~\cite{fan2008liblinear} on the tree-ensemble feature representation to solve the L2 regularized dual problem from equation~(\ref{eq:lr_dual}). For SVMs, we use the SVC solver from liblinear~\cite{pedregosa2011scikit} to solve equation~(\ref{eq:svm_dual}). Experiments are run on an Intel(R) Xeon(R) CPU E5-2690 v4 @ 2.6GHz with 30GB of RAM. Source code for TREX and all experiments is available at \url{https://github.com/jjbrophy47/trex}.

\paragraph{Hyperparameter Tuning}

In our experiments, we measure predictive performance using accuracy, and select the hyperparameters of our tree ensemble by performing five-fold cross-validation; we tune the number of trees~(from 10 to 250) and the maximum depth of each tree~(from 3 to unlimited depth). To tune the surrogate model that approximates the tree ensemble, we randomly select 10\% of the training data as validation and select the surrogate model whose predictions have the highest Pearson correlation to the tree ensemble predictions on the validation data. For the KLR and SVM models, C is tuned between $[10^{-2}, 10^2]$; for the KNN model, $k$ is tuned between $[3, 61]$.

\subsection{Tree-Ensemble Performance}

First, we verify that tree ensembles are necessary to obtain good predictive performance on these datasets. In particular, if other methods that are simpler and easier to interpret perform just as well, then there is no need to explain a complex tree ensemble.

Table~\ref{tab:performance} shows the test set accuracy for each dataset, comparing gradient boosted decision trees (GBDTs) to logistic regression (LR), support vector machines (SVM) with a linear kernel~(Linear) and RBF kernel~(RBF), and k-nearest neighbors (KNN). GBDTs are consistently more accurate, justifying the use of a tree ensemble and the need for methods to explain them.

\begin{table}[h]
\caption{Test Accuracy of GBDT vs. Interpretable Models}
\label{tab:performance}
\centering
\begin{small}
\begin{tabular}{lrrrr}
\toprule
Model & Churn & Amazon & Adult & Census \\
\midrule
GBDT & \textbf{0.813} & \textbf{0.947} &
      \textbf{0.868} & \textbf{0.958} \\
LR           & 0.806 & 0.940 & 0.824 & 0.948 \\
SVM (Linear) & 0.806 & 0.940 & 0.822 & 0.946 \\
SVM (RBF)    & 0.759 & 0.941 & 0.764 & 0.938 \\
KNN          & 0.762 & 0.939 & 0.802 & 0.946 \\
\bottomrule
\end{tabular}
\end{small}
\end{table}

\subsection{Fidelity}

To generate explanations, TREX first trains a surrogate model that approximates the predictive behavior of a tree ensemble.  We compare the fidelity of TREX-KLR and TREX-SVM to a KNN model built using the tree-ensemble kernel, which we denote TEKNN. For this and all subsequent experiments, we use the LeafOutput tree-ensemble kernel for these models\footnote{We focus on the LeafOutput kernel because it provides better fidelity than LeafPath and more detail than TreeOutput.
Fidelity results using 
other
kernels are in \S\ref{sec:fidelity_appendix} of the Appendix.}. We see in Figure~\ref{fig:fidelity} that TREX is able to approximate the tree ensemble's predictions very accurately, better than TEKNN in all scenarios.

\begin{figure}
\includegraphics[width=0.475\textwidth]{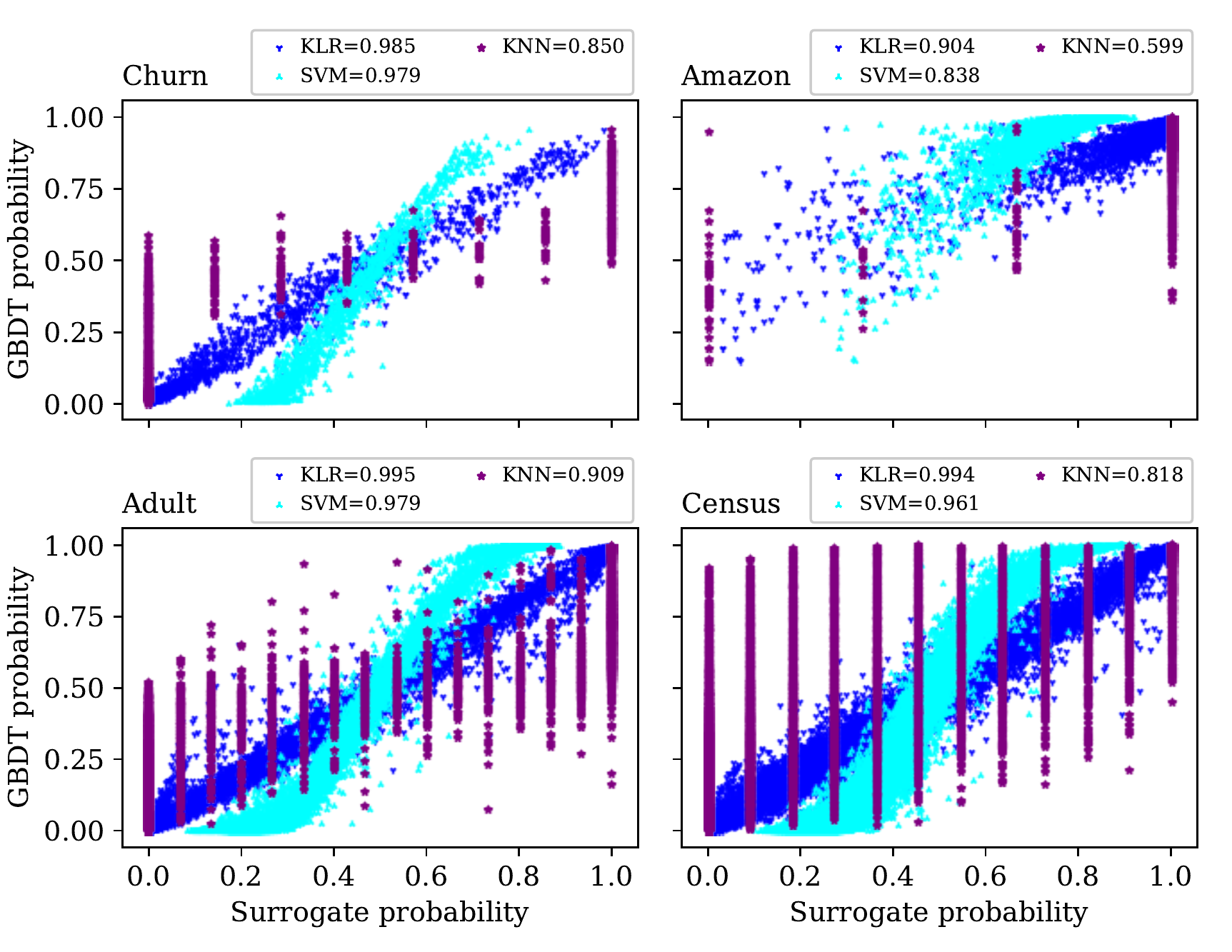}
\caption{Fidelity comparison between a tree ensemble and three surrogate models: two TREX models and one KNN model, all using the LeafOutput tree-ensemble kernel. The numbers in each plot represent the Pearson correlation between the GBDT and surrogate predictions. For the TREX-SVM model, we apply a sigmoid function to the output of its decision function to obtain predictions closer to probabilities for a closer comparison between all methods.}
\label{fig:fidelity}
\end{figure}

\subsection{Dataset Cleaning}\label{sec:cleaning}

\begin{figure*}
\includegraphics[width=\textwidth]{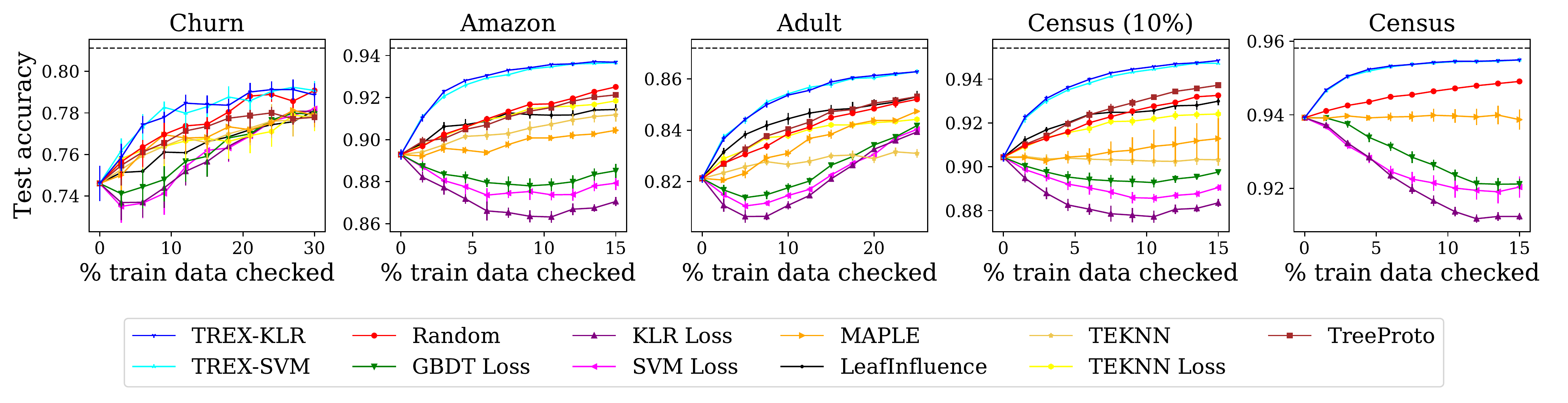}
\caption{Change in test accuracy as training points are checked and fixed; the dashed line represents test accuracy before label corruption. Each experiment is repeated 5 times to obtain standard error bars. \LeafInfluence{} and \emph{TEKNN} were too slow in generating predictions for the Census dataset, so we also include results for a 10\% subset of the Census dataset.}
\label{fig:cleaning}
\end{figure*}

Datasets often contain missing or noisy labels that can degrade the performance of a classifier. We can use TREX to efficiently identify problematic training instances and relabel them as appropriate.
Following a similar experimental setup as~\citet{koh2017understanding}, we corrupt a training set by randomly flipping 40\% of the training labels and training a tree ensemble model on the resulting noisy dataset. We then use TREX to order the training instances to be manually checked, fixing them if they had been previously flipped; the model is then evaluated on a held-out test set. When ordering the training samples to be checked, we take the same approach as \citet{yeh2018representer} and sort the training instances by the absolute value of their weights, $\alpha$.

We repeat each experiment five times, and compare our approach to the following baseline orderings: 
\begin{itemize}
\item \emph{Random}: A completely random ordering.

\item \emph{GBDT Loss}: An ordering based on the loss of the tree-ensemble predictions.

\item \emph{Surrogate Loss}: Orderings based on the loss of the surrogate models~(KLR, SVM, or KNN).

\item \emph{TEKNN}: An ordering based on how often each training point appears in each other's $k$-nearest neighbors.

\item \emph{MAPLE}~\cite{plumb2018model}:  Samples are ordered by similarity density.

\item \emph{LeafInfluence}~\cite{pmlr-v80-sharchilev18a}: An extension of influence functions to GBDT; samples are ordered by the influence of each training sample on itself~\cite{koh2017understanding}.

\item \emph{TreeProto}~\cite{tan2020tree}: A submodular prototype selection method designed specifically for tree ensembles.
\end{itemize}

Results are shown in Figure~\ref{fig:cleaning}. On every dataset, TREX-KLR and TREX-SVM achieve the highest accuracy for each amount of data checked. In other words, TREX identifies the training instances that (after relabeling) have the greatest impact on the model's test performance. In contrast, MAPLE, TEKNN, and several other baselines often perform \emph{worse than a random ordering}.

\subsection{Remove and Retrain}\label{sec:roar}

Inspired by a recent approach that measures explanation quality for feature attribution techniques, we adapt the ROAR (\textbf{R}em\textbf{O}ve \textbf{A}nd \textbf{R}etrain) framework~\cite{hooker2019benchmark} from measuring feature importance to measuring the influence of training samples on a set of model predictions. In this experiment, each method generates and aggregates a set of instance-attribution explanations for a randomly selected set of $n=50$ test instances, and orders the training data from most positively influential to most negatively influential. Then, the training data is removed in 10\% increments, in which the best explanatory methods cause the sharpest degradation in predictive performance. Each experiment is repeated 20 times.

\begin{figure}
\includegraphics[width=0.475\textwidth]{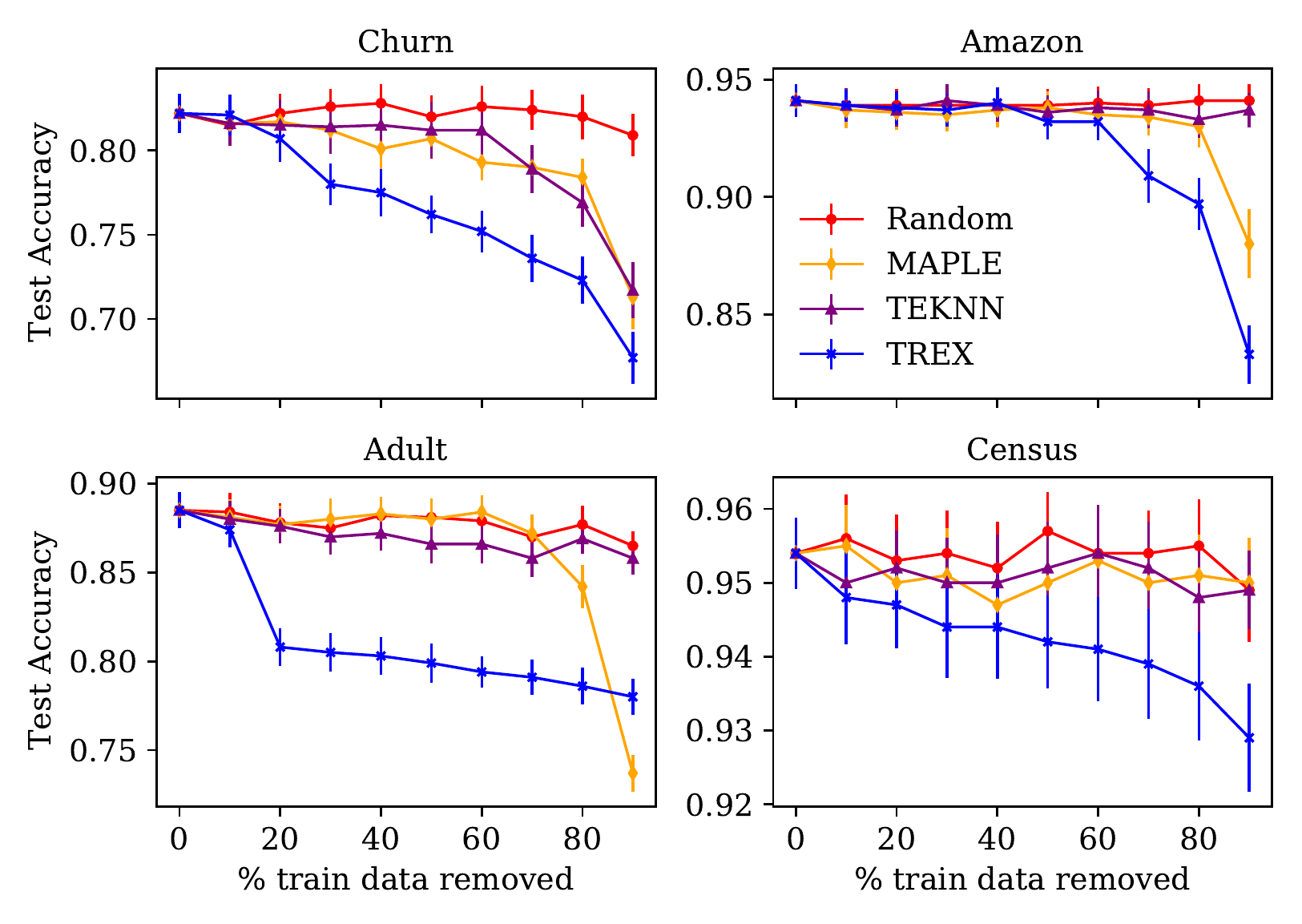}
\caption{Change in test performance as the training samples with the most positive influence on the test set are removed in 10\% increments up to 90\%.}
\label{fig:roar}
\end{figure}

We observe that each dataset is quite robust to the deletion of training instances, maintaining relatively high accuracy even as 90\% of the training data is randomly removed~(Figure~\ref{fig:roar}). However, we find that TREX\footnote{We use TREX-KLR for this experiment since the number of support vectors for TREX-SVM was typically a small fraction of the training set, resulting in a random ordering for most training samples.} is generally able to find the training instances that, when deleted, cause the earliest and largest degradation in model performance. For the highest amounts of data deletion, MAPLE shows a large degradation on the \texttt{Adult} dataset, but MAPLE and TEKNN are generally consistently worse than TREX; LeafInfluence was too inefficient to run (see timing experiments below) and TreeProto is a global explanation method, unable to provide the most influential training samples for a \emph{given} set of test predictions.

\subsection{Runtime Comparison}

In this section, we measure the time it takes for each method to generate a \emph{local} explanation for a single random test instance. We compare TREX to LeafInfluence~(to give LeafInfluence every advantage, we use the \emph{single point} update set, its fastest setting), MAPLE, and TEKNN.

\begin{table}
\caption{Time (in seconds) to compute the impact of all training instances on the prediction of a single test instance. MAPLE did not finish (DNF) training on the Census dataset after running for 12 hours, so the test time for that dataset is not applicable~(N/A). Each experiment is repeated 5 times to obtain average runtimes, but standard deviations are omitted for clarity and can be found in Table~\ref{tab:runtime_appendix} in \S~\ref{sec:runtime_appendix} of the appendix.}
\label{tab:runtime}
\begin{small}
\begin{tabular}{lrrrr}
\toprule
\textbf{Model} & \textbf{Churn} & \textbf{Amazon} & \textbf{Adult} & \textbf{Census} \\
\midrule
\multicolumn{5}{l}{\textbf{Training}} \\

LeafInf  & 0   & 0     & 0      & 0 \\
MAPLE    & 700 & 5,400 & 11,500 & DNF \\
TEKNN    & 20  & 1,000 & 500    & 25,000 \\
TREX-KLR & 20  & 190   & 90     & 1,900 \\
TREX-SVM & 20  & 190   & 90     & 1,900 \\

\midrule
\multicolumn{5}{l}{\textbf{Testing}} \\

LeafInf & 100\hphantom{.00} & 3,500\hphantom{.00} & 1,500\hphantom{.00} & 29,900\hphantom{.00} \\
MAPLE    & 0.03 & 0.05 & 0.06 & N/A \\
TEKNN    & 0.01 & 0.41 & 0.13 & 1.97 \\
TREX-KLR & 0.02 & 0.60 & 0.30  & 4.73 \\
TREX-SVM & 0.02 & 0.60 & 0.30  & 4.69 \\

\bottomrule
\end{tabular}
\end{small}
\end{table}

The measurement is broken up into training and testing costs. Training is a one-time cost to setup the explainer~(tuning any hyperparameters and training the explainer), while testing is the time it takes to use the explainer to compute the influence of each training sample on the prediction of the randomly chosen test instance. Each experiment is repeated five times; results are in Table~\ref{tab:runtime}. The LeafInfluence method has no setup cost since it does not train any type of surrogate model; however, the bulk of its time is spent computing the influence of each training instance, taking over 12 hours for the Census dataset~(Table~\ref{tab:runtime}). MAPLE has the largest setup cost, taking over three hours to train on the Adult dataset. TEKNN shows a slight advantage in testing cost compared to TREX, but does not scale as well in terms of training time.

TREX incurs a relatively modest one-time training cost to train the kernelized model, roughly 1-2 orders of magnitude faster than MAPLE. After this step, TREX is able to quickly generate an explanation for a test instance, roughly 3-4 orders of magnitude faster than LeafInfluence.
Overall, our approach takes relatively little time to train while also being fast enough to generate instance-attribution explanations in real time for individual queries, or even for groups of test instances, as shown in Section~\ref{sec:roar}.

\subsection{Understanding Misclassified Predictions}

To evaluate the utility of TREX in explaining individual predictions, we created a domain mismatch within the Adult dataset. In the original training set, all 395 people under the age of 18 are labeled as negative, indicating that they make less than or equal to \$50,000 per year. We reduced that set to 98 people and flipped 83 of the labels, so that 83 out of 98 17-year-olds in the training set are positive. This inevitably caused incorrect predictions in the test set, where 17-year-olds were predicted to have incomes over \$50,000 per year.

\begin{figure}[t]
\centering
\includegraphics[width=0.475\textwidth]{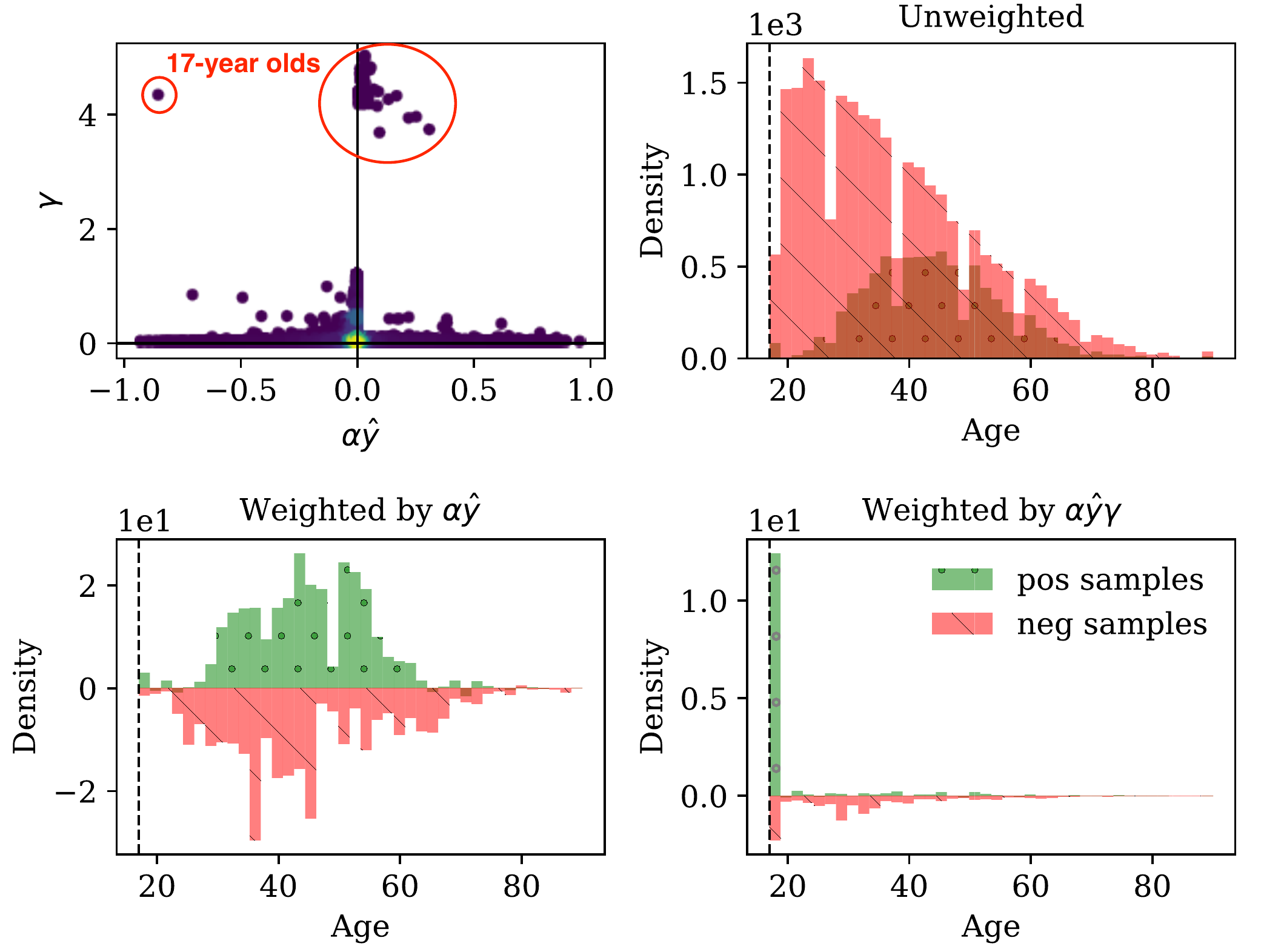}
\caption{Explanation for an incorrect test prediction of a 17-year old making $>50k$ per year in the Adult dataset. \emph{Top-left}: the similarity~($\gamma$) of each training sample to the test sample; the signed weight of each sample~($\alpha \hat{y}$) is also given as context. In this plot, we see a cluster of high similarity training points which contribute heavily towards the final test prediction. \emph{Top-right}: a histogram of the positive and negative-labeled training examples, shown for each age range. \emph{Bottom-left}: the same as top-right, except each training sample is weighted by its signed representer value~($\alpha \hat{y}$). \emph{Bottom-right}: the same as top-right, except each training sample is weighted by its contribution~($\alpha \hat{y} \gamma$) to the test prediction.}
\label{fig:dataset_shift}
\end{figure}

We then selected one of these incorrect predictions and used TREX-KLR\footnote{We see similar results with TREX-SVM, but focused on a smaller number of points due to the sparse SVM solution.} to help explain it. After computing the influence of each training sample on the prediction, we first plotted the similarity of each training sample to the test sample~($\gamma$) as well as the signed weight of each training sample~($\alpha \hat{y}$)~(Figure~\ref{fig:dataset_shift}: top-left). We immediately notice a clump of high-similarity points, along with a few other outliers; these high similarity points~($\gamma > 2.0$) have a high impact on the test prediction and make up 97 out of the 98 17-year-olds in the training set.

Investigating further, we show the distribution of the age attribute in three histograms. In the training set~(top-right), most examples with low age ($< 25$) are negative~(i.e. samples that do not have a positive label). After weighting by $\alpha \hat{y}$ (bottom-left), their contribution to the overall model is small but positive.
However, for this one test example (bottom-right), these same points make a very large, positive contribution~($\alpha_i \hat{y}_i \gamma_i$ is the contribution of training sample $i$ to the test prediction), explaining the final prediction of a positive label. Thus, we see that these incorrectly labeled 17-year-olds in the training set are having a very large impact on the incorrect prediction of this 17-year-old in the test set.

\section{Conclusion}

In this work we have developed TREX, a method of explaining tree ensemble predictions via the training data. We extended the representer point framework~\cite{yeh2018representer} to work for non-differentiable tree ensembles by exploiting the tree-ensemble structure to create a new tree-based kernel, from which we can train a kernelized model. 

We demonstrated that this model is capable of closely approximating the predictive behavior of the tree ensemble, and can be used to aid in dataset debugging better understand model behavior; TREX is also significantly faster than alternative methods in terms of setup and explanation costs.

\section{Discussion}

A future direction could include an in-depth investigation into the robustness of TREX and other instance-attribution methods to adversarial perturbations; having robust explanations is especially useful if we want predictive modeling to become more widely adopted in certain domains~(e.g. medical).

Also, since TREX provides explanations from the perspective of influential training samples, it can be combined with other explanation methods such as feature-attribution explanations to provide the most comprehensive view of all the elements that contribute towards a certain prediction.

Overall, understanding how individual predictions are made can affect all levels of the machine learning pipeline. ML practitioners can benefit from explanations to help them debug their models; business executives can make more informed decisions about model deployment based on when and why their models fail, and consumers of these models may need explanations for specific decisions, especially if those decisions can significantly impact their life~(e.g. bank loan). We believe our work contributes to these goals by shedding more light into the inner workings of tree-ensembles.

\bibliography{references}

\newpage

\onecolumn
\appendix

\section{Experiment Details}

\subsection{Standard Deviations for Runtime Comparison}\label{sec:runtime_appendix}

This section presents detailed results for the runtime comparison experiment; specifically, the runtimes with standard deviations~(Table~\ref{tab:runtime_appendix}).

\begin{table*}[h]
\centering
\caption{Time (in seconds) to compute the impact of all training instances on a single test instance. MAPLE did not finish (DNF) training on the Census dataset after running for 12 hours, so the test time for that dataset is not applicable~(N/A). Each experiment is repeated 5 times to obtain average runtimes and standard deviations (S.D.).}
\label{tab:runtime_appendix}
\begin{small}
\begin{tabular}{lrrrrrrrr}
\toprule
& \multicolumn{2}{c}{\textbf{Churn}} & \multicolumn{2}{c}{\textbf{Amazon}} & \multicolumn{2}{c}{\textbf{Adult}} & \multicolumn{2}{c}{\textbf{Census}} \\
\textbf{Model} & Average & S.D. & Average & S.D. & Average & S.D. & Average & S.D. \\
\midrule
\multicolumn{5}{l}{\textbf{Training}} \\

LeafInf  & 0 & (N/A)  & 0 & (N/A)     & 0 & (N/A)    & 0 & (N/A) \\
MAPLE    & 700 & (10\hphantom{.0})    & 5,400 & (70) & 11,500 & (80) & DNF & (N/A) \\
TEKNN    & 20 & (0.2)  & 1,000 & (20) & 500 & (10)   & 25,000 & (1,500) \\
TREX-KLR & 20 & (0.5)  & 190   & (1)  & 90 & (1)     & 1,900 & (40) \\
TREX-SVM & 20 & (0.2)  & 190   & (3)  & 90 & (1)     & 1,900 & (40) \\

\midrule
\multicolumn{5}{l}{\textbf{Testing}} \\

LeafInf & 100\hphantom{.00} & (0.6\hphantom{00}) &
          3,500\hphantom{.00} & (12\hphantom{.000}) &
          1,500\hphantom{.00} & (10\hphantom{.000}) &
          29,900\hphantom{.00} & (600\hphantom{.00}) \\
MAPLE    & 0.03 & (0.001) & 0.05 & (0.003) & 0.06 & (0.005) & N/A & (N/A) \\
TEKNN    & 0.01 & (0.001) & 0.41 & (0.041) & 0.13 & (0.020) & 1.97 & (0.45) \\
TREX-KLR & 0.02 & (0.000) & 0.60 & (0.007) & 0.30 & (0.005)  & 4.73 & (0.07) \\
TREX-SVM & 0.02 & (0.000) & 0.60 & (0.003) & 0.30 & (0.001)  & 4.69 & (0.05) \\

\bottomrule
\end{tabular}
\end{small}
\end{table*}

\subsection{Tree-Ensemble Hyperparameters}

We tune the number of trees in the GBDT model using values [10, 100, 250], and the maximum depth of each tree using values [3, 5, 10, unlimited]. Chosen hyperparameter values of the GBDT model for each dataset are in Table~\ref{tab:gbdt_hyperparameters}.

\begin{table*}[h]
\centering
\caption{Hyperparameters used for the CatBoost GBDT model.}
\label{tab:gbdt_hyperparameters}
\begin{small}
\begin{tabular}{lrr}
\toprule
\textbf{Dataset} & \textbf{No. Trees} & \textbf{Max Depth} \\
\midrule
Churn & 100 & 3 \\
Amazon & 250 & 5 \\
Adult & 100 & 5 \\
Census & 250 & 5 \\
\bottomrule
\end{tabular}
\end{small}
\end{table*}

\section{Additional Experiments}

\subsection{Fidelity Using Different Tree-Ensemble Kernels}\label{sec:fidelity_appendix}

\begin{figure*}
  \centering
  \begin{minipage}[b]{0.48\textwidth}
    \includegraphics[width=\textwidth]{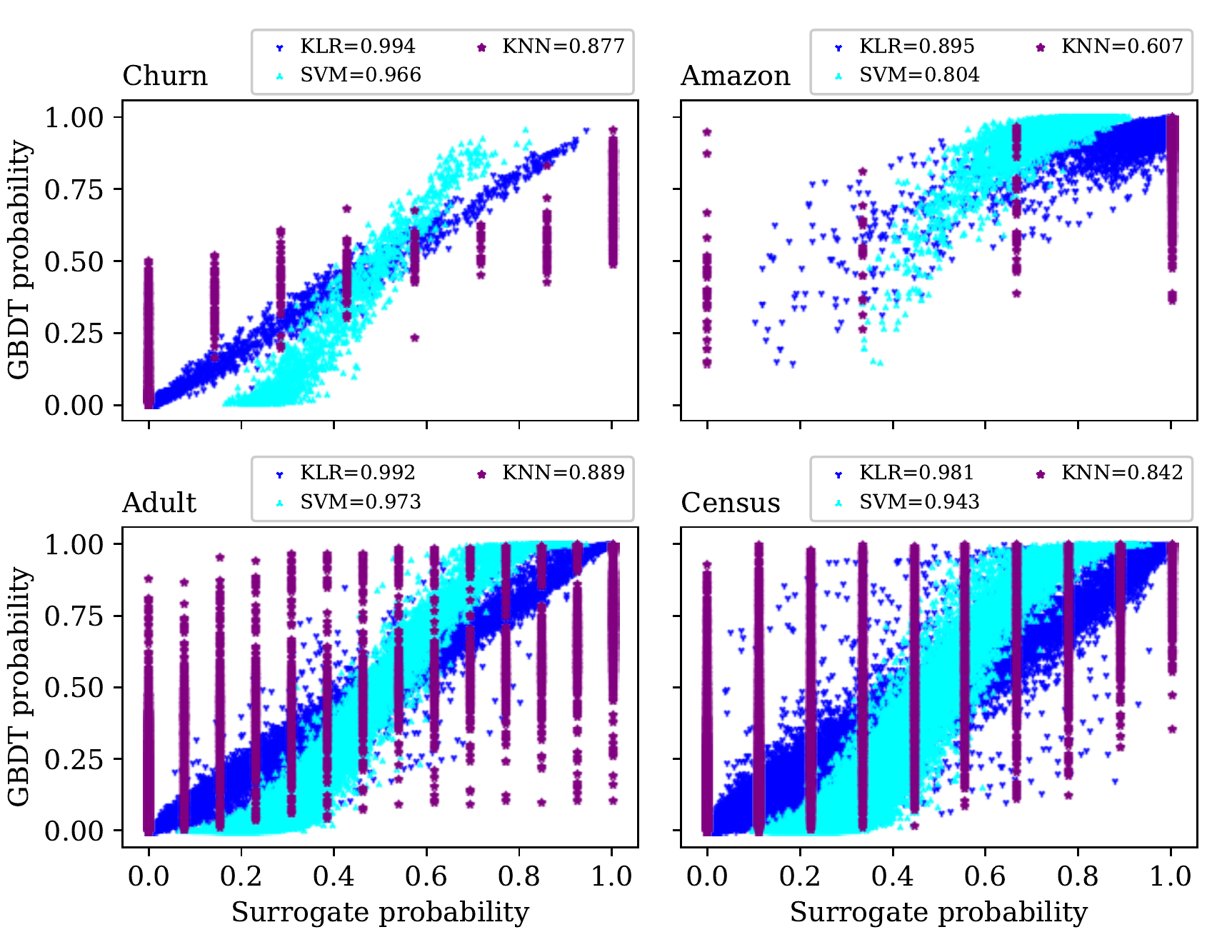}
    \caption{Fidelity results using the LeafPath tree-ensemble kernel.}
    \label{fig:fidelity_leaf_path}
  \end{minipage}
  \hfill
  \begin{minipage}[b]{0.48\textwidth}
    \includegraphics[width=\textwidth]{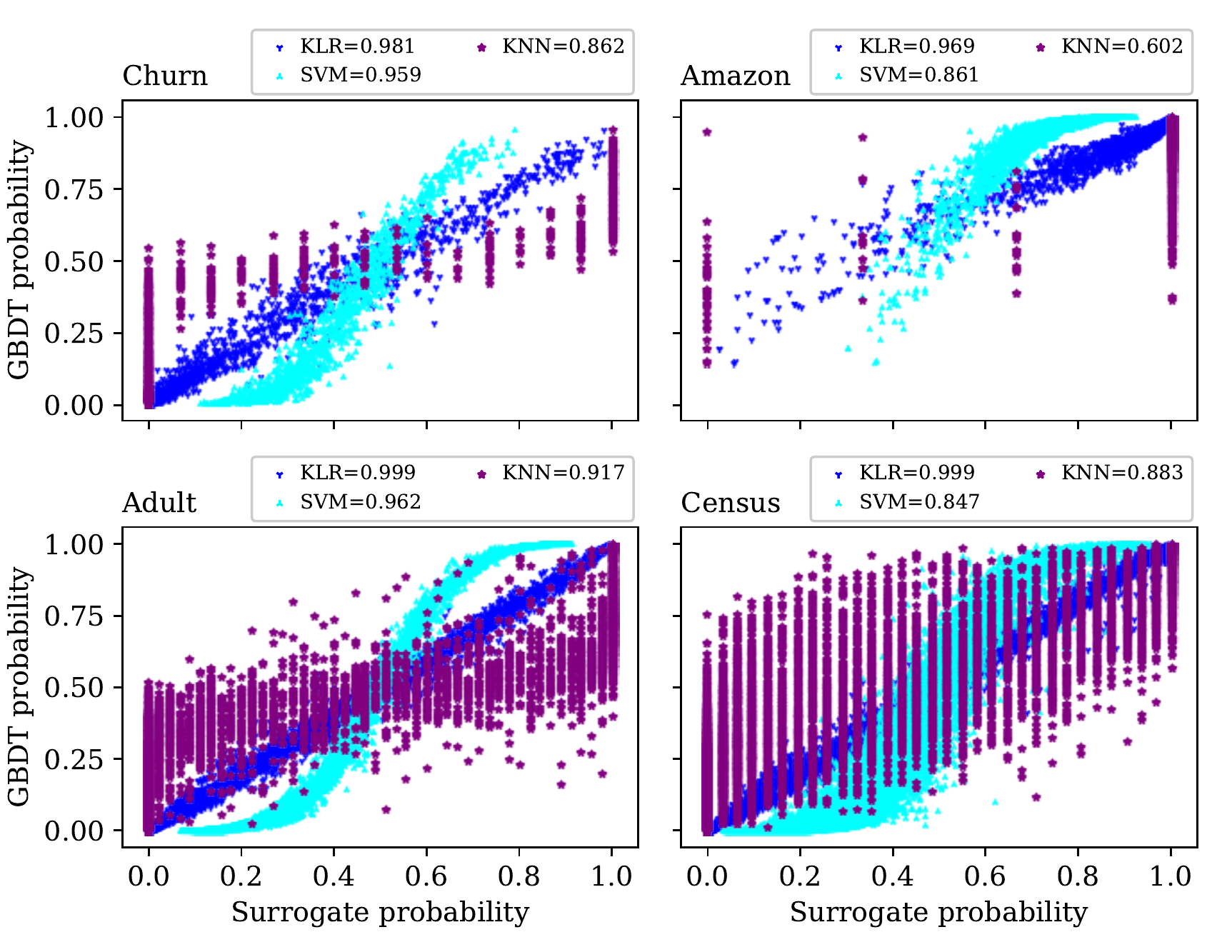}
    \caption{Fidelity results using the TreeOutput tree-ensemble kernel.}
    \label{fig:fidelity_tree_output}
  \end{minipage}
\end{figure*}

Figures~\ref{fig:fidelity_leaf_path}~and~\ref{fig:fidelity_tree_output} showfidelity results using the LeafPath and TreeOutput tree-ensemble kernels.

\end{document}